\documentclass[runningheads]{llncs}
\usepackage[T1]{fontenc}
\usepackage{graphicx}
\usepackage{multirow}
\usepackage{booktabs}
\usepackage{adjustbox}
\usepackage[mathscr]{euscript}
\DeclareSymbolFont{rsfs}{U}{rsfs}{m}{n}
\DeclareSymbolFontAlphabet{\mathscrsfs}{rsfs}
\usepackage{csquotes}
\usepackage{algorithmic}
\usepackage{csquotes}
\usepackage{cite}
\usepackage{amsmath,amssymb,amsfonts}
\usepackage{algorithmic}
\usepackage{graphicx}
\usepackage{textcomp}

\usepackage{bbold}
\usepackage{caption}

\usepackage{float}
\usepackage{graphicx}
\usepackage{caption}
\usepackage{array}
 \usepackage{subcaption}
 \usepackage{comment}
 \usepackage{rotating}
\usepackage{pgf}

\begin{document}
%
%\title{Random Weighted Perturbation and Contrastive Learning based Meta Predictor for Health Mention Classification}
\title{A Unique Training Strategy to Enhance Language Models Capabilities for Health Mention Detection from Social Media Content}
\titlerunning{Meta Predictor for Health Mention Classification}

\author{Pervaiz Iqbal Khan\inst{1,2}\orcidID{0000-0002-1805-335X} \and
Muhammad Nabeel Asim\inst{1}\orcidID{0000-0001-5507-198X}
\and
Andreas Dengel\inst{1,2}\orcidID{0000-0002-6100-8255} \and
Sheraz Ahmed\inst{1}\orcidID{0000-0002-4239-6520}}
\authorrunning{Pervaiz Iqbal Khan et al.}
\institute{
German Research Center for Artificial Intelligence (DFKI), Kaiserslautern, Germany \and
RPTU Kaiserslautern-Landau  \\
\email{\{pervaiz.khan,muhammad\_nabeel.asim,andreas.dengel,sheraz.ahmed\}@dfki.de}\
\email{}
\
}

\maketitle              % typeset the header of the contribution
\begin{abstract}
An ever-increasing amount of social media content requires advanced AI-based computer programs capable of extracting useful information. Specifically, the extraction of health-related content from social media is useful for the development of diverse types of applications including disease spread, mortality rate prediction, and finding the impact of diverse types of drugs on diverse types of diseases. Language models are competent in extracting the syntactic and semantics of text. However, they face a hard time extracting similar patterns from social media texts. The primary reason for this shortfall lies in the non-standardized writing style commonly employed by social media users. Following the need for an optimal language model competent in extracting useful patterns from social media text, the key goal of this paper is to train language models in such a way that they learn to derive generalized patterns. The key goal is achieved through the incorporation of random weighted perturbation and contrastive learning strategies. On top of a unique training strategy, a meta predictor is proposed that reaps the benefits of 5 different language models for discriminating posts of social media text into non-health and health-related classes. Comprehensive experimentation across 3 public benchmark datasets reveals that the proposed training strategy improves the performance of the language models up to $3.87\%$, in terms of F1-score, as compared to their performance with traditional training. Furthermore, the proposed meta predictor outperforms existing health mention classification predictors across all 3 benchmark datasets.
%\keywords{Health mention classification  \and Random weighted perturbation \and Contrastive learning.}

\end{abstract}

\section{Introduction}
Social media platforms like Facebook\footnote{https://www.facebook.com/}, Twitter\footnote{https://twitter.com}, and Reddit\footnote{https://www.reddit.com/} have played a significant role in connecting people worldwide, effectively shrinking the world into a global community \cite{abbas2019impact}. Through these platforms, people can communicate and share their opinions regarding product quality, service standards, and even their emotions and concerns about health issues. Various brands are utilizing the power of artificial intelligence (AI) methods and social media platforms for enhancing the quality of their products and services. To empower the process of brand monitoring by analyzing social media content, there is a marathon for developing more robust and precise deep learning predictors\cite{wassan2021amazon, jagdale2019sentiment,ray2017twitter,fang2015sentiment}. Similar to brand monitoring, healthcare systems can be improved by closely monitoring social media content; however, little attention has been paid in this regard and a few predictors have  been proposed for discriminating health-related content from other conversations. Accurate categorization of social media content into health and general discussion related classes can provide significant insights about the spread of diseases, mortality rate analysis, the impact of drugs on individuals \cite{huang2022predicting,jahanbin2020using}, etc.

Among diverse types of predictors BERT \cite{devlin2018bert}, RoBERTa \cite{liu2019roberta} and XLNet \cite{yang2019xlnet}, etc. are producing state-of-the-art (SotA) performance values for diverse types of text classification tasks. The prime reason behind their better performance is the utilization of transfer learning that helps them understand the contextual information of content. Trained language models offer better performance on small datasets where they are trained in a supervised fashion. These language models are also being utilized for the development of diverse types of real-world useful applications such as DNA \cite{zeng2023mulan, mock2022taxonomic}, RNA sequence classification \cite{yang2022scbert}, Fake URL detection \cite{maneriker2021urltran}, hate speech detection \cite{mozafari2020bert}, fake news detection \cite{tariq2022adversarial}, and brand monitoring\cite{jagdale2019sentiment, ray2017twitter}. Researchers are trying to utilize these language models to strengthen healthcare systems by developing diverse types of applications including mortality rate prediction, disease spread prediction \cite{li2021long}, and disease symptoms prediction \cite{luo2021deep}. Development of these applications requires discrimination of non-health mention related social media content from health mention content. Precise discrimination of this content is difficult as compared to the aforementioned tasks as people use non-standardized and abbreviated forms of words while describing diseases, symptoms, and drugs. Furthermore, the posts indicating the presence of disease and general conversation about the diseases may contain similar words, hence learning the appropriate context becomes more non-trivial. This paper proposes a random weighted perturbation (RWP) method for finetuning language models on the HMC task, where it generates perturbations based on the statistical distribution of the model's parameters. Furthermore, it jointly optimizes language model training by taking both perturbed and unperturbed parameters. Moreover, it introduces contrastive learning as an additional objective function to further improve text representation and the classifier's performance.

Manifold contributions of this paper can be summarized as:
\begin{itemize}
    \item It presents a unique strategy to train language models that avoids over-fitting and improves their generalization capabilities.
    \item Over 3 public benchmark datasets, it performs large-scale experimentation, where the aim is to highlight the impact of the proposed training strategy over diverse types of language models. 
    \item With an aim to develop a robust and precise predictor for health mention classification, it presents a meta predictor that reaps the benefits of 5 different language models.
    \item Combinely, the proposed meta predictor along with the unique training strategy produces SotA performance values over 3 public benchmark datasets.
    
\end{itemize}
\section{Related Work}
This section presents 11 different types of predictors that are proposed for the HMC task. The presented predictors employ different approaches, such as convolutional neural networks (CNNs), language models, and hybrid approaches.

Iyer et al.  \cite{iyer2019figurative} proposed a two-step approach in which binary features were extracted to indicate whether a disease word was used figuratively. This information was passed to the CNN model as an additional feature for tweet classification. Luo et al. \cite{luo2022covid} presented a health mention classifier that used dual CNN to find COVID-19 mentions. The dual CNN had two components: an additional network called an auxiliary network (A-Net), which helped the primary network (P-Net) overcome the class imbalance. 

Karisani et al. \cite{karisani2018did} proposed WESPAD, which divided and distorted the embedding space to make the model generalize better on examples that had not been seen. Jiang et al. \cite{jiang2018identifying} used non-contextual embeddings to extract representations of the tweets, and then passed those representations to Long Short-Term Memory Networks \cite{hochreiter1997long}. By adding 14k new tweets related to 10 diseases, Biddle et al. \cite{biddle2020leveraging} extended the PHM-2017 \cite{karisani2018did} dataset. Moreover, they incorporated sentiment information with contextual and non-contextual embeddings into their study. The incorporation of sentiment information significantly enhanced the classification performance of the classification algorithms. Khan et al. \cite{khan2020improving} improved the performance of the health mention classifier by incorporating emojis into tweet text and using XLNet-based \cite{yang2019xlnet} word representations. Naseem et al. \cite{naseem2022identification} presented a new health mention dataset related to Reddit posts, and classified the posts based on a combination of disease or symptom terms and user behavior. A method for learning the context and semantics of HMC was proposed by Naseem et al. \cite{naseem2022robust}. The method used domain-specific word representations and Bi-LSTM with an attention module. Naseem et al. \cite{naseem2022benchmarking} proposed the PHS-BERT, a domain-specific pretrained language model (PLM) for social media posts, and evaluated the model on various social media datasets. However, pre-training language models requires a significant amount of time and computing resources.

A contrastive adversarial training method was presented by Khan et al. \cite{khan2022improving} using the Fast Gradient Sign Method (FGSM) \cite{goodfellow2014explaining} to perturb the embedding matrix parameters of the model. Then, they jointly trained the clean and adversarial examples, and employed the contrastive learning. Khan et al. \cite{khan2022novel} presented a training method for HMC, in which they added Gaussian noise to the hidden representations of language models. The cleaned and perturbed examples were trained simultaneously. The empirical evidence demonstrated that perturbing the earlier layers of the model enhanced its performance. This paper proposes a new training strategy for language models that uses random weighted perturbations in the parameters of language models and additionally employs contrastive learning to improve text representations produced by the models.
\section{Method}
This section briefly describes the overall workflow of the proposed training strategy and the  meta predictor built based on that strategy. Moreover, it introduces the basic components of the proposed method, i.e., transformers, random weighted perturbations and contrastive learning.
\subsection{Proposed Training Method and Meta Predictor}
\begin{figure}
\includegraphics[width=\textwidth]{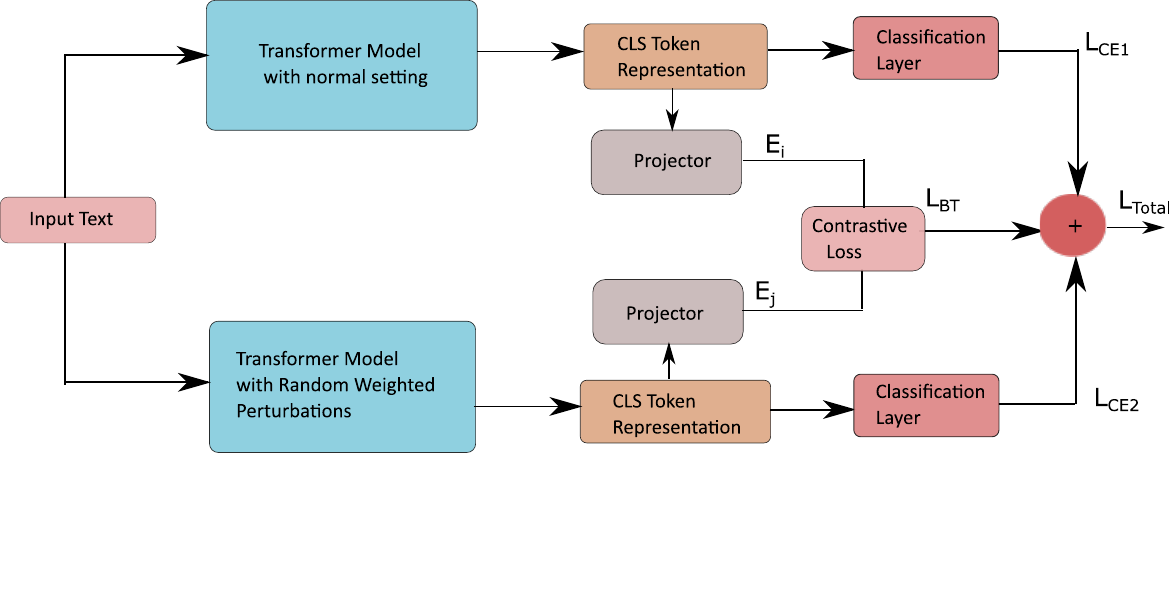}
\caption{Graphical illustration of the proposed strategy for language models training.} \label{fig1}
\end{figure}

Fig. \ref{fig1} illustrates the workflow of the proposed training strategy, where it can be seen that textual data is processed at two different streams of transformer models, i.e., one stream with normal transformer model setting, and the other with random weighted perturbations. Here, transformer models represent 5 different language models, namely, BERT \cite{devlin2018bert}, RoBERTa \cite{liu2019roberta}, DeBERTa \cite{he2020deberta}, XLNet \cite{yang2019xlnet}, and GPT-2 \cite{radford2018improving}. Furthermore, for both streams, loss is computed using cross-entropy. Moreover, an additional third stream is introduced, where CLS token representations are extracted from both the original and perturbed models for each training example and then projected to lower dimensions using a projection network. Then contrastive loss is computed using the Barlow Twins \cite{zbontar2021barlow} method that takes these two lower dimensional representations as inputs. Finally, the total loss $\mathcal{L}_{total}$ is computed as follows:
\begin{equation}
\mathcal{L}_{total} = \frac{(1- \lambda )}{2} (\mathcal{L}_{CE_1} + \mathcal{L}_{CE_2}) + \lambda \mathcal{L}_{BT}
\end{equation}
where `$\mathcal{L}_{BT}$' is the Barlow Twins \cite{zbontar2021barlow} loss, `$\mathcal{L}_{CE_1}$', and `$\mathcal{L}_{CE_2}$' are two cross-entropy losses, and `$\lambda$' is the hyperparameter that controls the weight between the three losses.

To reap the benefits of various transformer models and improve the prediction score, a meta predictor is proposed at the test time. Specifically, the class prediction probabilities of each of the 5 models are taken for every test-set post, and then averaged. Then the maximum probability for each post is treated as the predicted class. Results suggest that an ensemble of model predictions improves the overall classification scores for all the datasets.

A more comprehensive detail about different modules of the proposed method is presented in the following subsections:

\subsection{Transformers}
Let $\mathscrsfs{D} = \{x_i, y_i\}_{i=1,..., N}$ be a dataset having `N' examples with `X' data points and their corresponding `Y' labels. Let `\textbf{M}' be a pre-trained language model such as BERT \cite{devlin2018bert}, RoBERTa \cite{liu2019roberta}, DeBERTa \cite{he2020deberta}, etc. Each training example is represented as `T' tokens, i.e., $x_i = \{CLS, t_1, t_2,,...,t_T, SEP\}$, where `CLS', and `SEP' are special tokens. This token representation is passed as an input to `\textbf{M}' that produces the embeddings, i.e., $\{h^{L}_{CLS}, h^{L}_{1},...., h^{L}_{T},h^{L}_{SEP}\}$ for the given input example. Here, `L' represents the total layers in model `M'. To finetune `M', a classification layer is added on top of `M', and then the cross-entropy loss is minimized as follows:
 \begin{equation}\label{eq:loss}
\mathcal{L}_{CE} = - {\dfrac{1}{N}}\sum_{i=1}^{N}\sum_{c=1}^{C}y_{i,c}log(p(y_i,c|h^{i}_{[CLS]}))
    \end{equation}
 Here, `C' represents the total no. of classes, and $h_{[CLS]}$ contains the representations of the complete example.
\subsection{Random Weighted Perturbations and Contrastive Learning}
Let $w=\{w_1, w_2,...., w_L\}$ be the model parameters in all the layers. For each model parameter $w_L$, perturbations are generated following the Gaussian distribution $N(0,\epsilon \lVert w_L \rVert_{2})$, elementally, where $\epsilon$ controls the strength of perturbations, and $\lVert w_L \rVert_{2}$ is $L_2$ norm. The parameters with higher norm values will have a higher perturbation amount. The generated perturbations are added into the model's parameters. 

Contrastive Learning (CL) methods take two inputs, one of which is clean and the other is its perturbed version. The objective of CL is to learn embeddings for the inputs such that both the clean and its perturbed versions are pushed close, and other examples are pushed apart in the representation space. Barlow Twins \cite{zbontar2021barlow} (BT) is a CL method that works on the redundancy reduction principle. Instead of perturbing the data point itself, in our experiments, the output of the perturbed model is treated as a perturbed input to the BT. Let $E^{c}$ be the embedding of the clean example and $E^{p}$ be the embedding of its perturbed version.  Then, the objective function of the Barlow Twins \cite{zbontar2021barlow} is minimized as follows:
\begin{equation}
 \mathcal{L_{BT}} = \sum_{i=1} (1 - A_{ii})^2 + \beta \sum_{i=1} \sum_{j \neq i}A_{ij}^2
\end{equation}
where $\sum_{i=1} (1 - A_{ii})^2$, and $\sum_{i=1} \sum_{j \neq i}A_{ij}^2$ represent invariance, and redundancy reduction terms respectively, and `$\beta$' is the trade-off parameter controlling weights between the two terms. The matrix `A' computes the cross-correlation between $E^{c}$, and $E^{p}$. `A' is computed as follows:

\begin{equation}
    A_{ij} = \frac{\sum_{b=1}^{N} E^{c}_{ b, i} E_{b, i}^{p} }{\sqrt{ \sum_{b=1}^{N} (E^{c}_{b, i} })^2    \sqrt{ \sum_{b=1}^{N} (E^{p}_{b, i} })^2 }
\end{equation}
where `b' is the batch size, and $A_{ij}$ represents the entry of the i-th row and j-th column of the matrix A. 
\section{Experiments}
In this section, we present the benchmark datasets used for training and evaluating our proposed method. We then present the training details used in the experiments.
\subsection{Benchmark Datasets}
The selection of appropriate data is an essential part of evaluating the predictive performance of a proposed predictor. Following the experimental settings and evaluation criteria of existing studies \cite{khan2022improving}, we evaluated our proposed predictor using 3 different public benchmark datasets. A comprehensive description of the development process of these datasets is provided in existing studies \cite{khan2022improving}, so here we only summarize the statistics of these datasets as shown in Table \ref{tbl:stats}.
\begingroup
\setlength{\tabcolsep}{2pt} % Default value: 6pt
\renewcommand{\arraystretch}{1.3}
\begin{table*}[!htbp]\centering
\caption{Statistics of the benchmark datasets used for validating the proposed training strategy.}
\label{tbl:stats}
\begin{tabular}{llll}
\toprule
\hline
\multicolumn{1}{c}{Dataset}      & \multicolumn{1}{c}{No. of Samples} & \multicolumn{1}{c}{No .of Diseases} & \multicolumn{1}{c}{No. of Classes} \\
\hline
\multicolumn{1}{c}{HMC-2019 \cite{biddle2020leveraging}}     & \multicolumn{1}{c}{15,742}        & \multicolumn{1}{c}{10}              & \multicolumn{1}{c}{2}              \\
\hline
\multicolumn{1}{c}{RHMD \cite{naseem2022identification}}        & \multicolumn{1}{c}{10,015}        & \multicolumn{1}{c}{15}              & \multicolumn{1}{c}{3}              \\

\hline
\multicolumn{1}{c}{COVID-19 PHM \cite{luo2022covid}} & \multicolumn{1}{c}{9,219}         & \multicolumn{1}{c}{1}               & \multicolumn{1}{c}{2}         
\\
\hline
\bottomrule
\end{tabular}
\end{table*}
\endgroup

Furthermore, we performed preprocessing over these datasets and removed user mentions, hashtags, and URLs from the COVID-19 PHM \cite{luo2022covid} and HMC-2019 \cite{biddle2020leveraging} datasets. Following the existing studies \cite{khan2020improving, khan2022improving}, we converted the emojis in the tweet into text, and  removed special characters such as \enquote{: and -}  and used those textual representations as a part of the tweet. Moreover, we used maximum sequence length of 64, 68, and 215 for HMC-2019 \cite{biddle2020leveraging}, COVID-19 PHM \cite{luo2022covid}, and RHMD \cite{naseem2022identification} datasets, respectively.
\subsection{Training Details}
We used large versions of 5 different pre-trained language models, namely RoBERTa \cite{liu2019roberta}, BERT \cite{devlin2018bert}, DeBERTa \cite{he2020deberta}, XLNet \cite{yang2019xlnet}, and GPT-2 \cite{radford2018improving}, and finetuned them for the 3 datasets. We used AdamW \cite{loshchilov2018fixing} as an optimizer for all  experiments. The selection of optimal parameters is regarded as an essential step in achieving superior performance. To obtain optimal values of hyperparameters, we tweaked several hyperparameters such as batch size b $\in \{16, 32\}$, $\epsilon \in \{5e^{-4}, 1e^{-4}, 5e^{-3}, 1e^{-3}\}$, $\lambda \in \{0.1, 0.2, 0.3, 0.4, 0.5\}$, and a learning rate of $1e^{-5}$ using grid-search. Furthermore, we trained all models for $40$ epochs with an early stopping strategy based on the validation set F1-score. To compute contrastive loss, we first projected the original representations of the CLS token to lower dimensions of 300 using a projection network and then computed contrastive loss. The projection network consists of two linear layers with a hidden unit size of $1024$, and an output layer with a hidden unit size of $300$, respectively. As an activation function, we used ReLU and applied 1-D batch normalization between the two linear layers. For BT \cite{zbontar2021barlow} loss, we used the default hyperparameters.
\section{Results and Analysis}
In this section, we present the results of the proposed training strategy and the traditional training strategy. Furthermore, we compare the proposed meta predictor results with existing methods in the literature.
\subsection{Contrastive Learning and  Random Weighted Perturbations Impact on the Predictive Performance of different Language Models}
Table \ref{tbl:main} shows the results of 5 different language models with traditional training strategy and proposed strategy.
Overall, 4 models out of 5 performed better with the addition of perturbation and contrastive learning on all 3 datasets in terms of F1 as an evaluation measure. On the contrary, one model, DeBERTa \cite{he2020deberta}, showed different behavior.
\begingroup
\setlength{\tabcolsep}{2pt} % Default value: 6pt
\renewcommand{\arraystretch}{1.3}
\begin{table*}[!htbp]\centering
\caption{showing that random weighted perturbation with contrastive learning (RWP + CL) improves F1-scores (macro) on 3 HMC datasets. Results are reported on the test sets of all the datasets.}
\label{tbl:main}
\begin{tabular}{llll}
\toprule
\hline
\multicolumn{1}{c}{Model}              & \multicolumn{1}{c}{HMC-2019 \cite{biddle2020leveraging}}       & \multicolumn{1}{c}{RHMD \cite{naseem2022identification}}           & \multicolumn{1}{c}{COVID-19 PHM \cite{luo2022covid}}   \\
\hline
\bottomrule
\multicolumn{1}{c}{BERT baseline  \cite{devlin2018bert}}      & \multicolumn{1}{c}{92.00}          & \multicolumn{1}{c}{79.81}          & \multicolumn{1}{c}{78.22}          \\
\hline
\multicolumn{1}{c}{BERT \cite{devlin2018bert} + RWP +CL}     & \multicolumn{1}{c}{\textbf{92.47}} & \multicolumn{1}{c}{\textbf{80.30}} & \multicolumn{1}{c}{\textbf{79.65}} \\
\hline
\multicolumn{1}{c}{RoBERTa baseline \cite{liu2019roberta}}   & \multicolumn{1}{c}{93.58}          & \multicolumn{1}{c}{80.20}          & \multicolumn{1}{c}{76.80}          \\
\hline
\multicolumn{1}{c}{RoBERTa \cite{liu2019roberta} + RWP + CL} & \multicolumn{1}{c}{\textbf{93.95}} & \multicolumn{1}{c}{\textbf{80.72}} & \multicolumn{1}{c}{\textbf{78.65}} \\
\hline
\multicolumn{1}{c}{DeBERTa baseline \cite{he2020deberta}}   & \multicolumn{1}{c}{93.66}          & \multicolumn{1}{c}{81.03}          & \multicolumn{1}{c}{\textbf{79.91}} \\
\hline
\multicolumn{1}{c}{DeBERTa \cite{he2020deberta} + RWP + CL} & \multicolumn{1}{c}{\textbf{93.91}} & \multicolumn{1}{c}{\textbf{82.36}} & \multicolumn{1}{c}{78.93}          \\
\hline
\multicolumn{1}{c}{XLNet baseline \cite{yang2019xlnet}}     & \multicolumn{1}{c}{92.53}          & \multicolumn{1}{c}{79.91}          & \multicolumn{1}{c}{77.43}          \\
\hline
\multicolumn{1}{c}{XLNet \cite{yang2019xlnet
} + RWP + CL}  & \multicolumn{1}{c}{\textbf{93.28}} & \multicolumn{1}{c}{\textbf{81.21}} & \multicolumn{1}{c}{\textbf{79.02}} \\
\hline
\multicolumn{1}{c}{GPT-2 baseline \cite{radford2018improving}}     & \multicolumn{1}{c}{88.05}          & \multicolumn{1}{c}{71.63}          & \multicolumn{1}{c}{71.22}          \\
\hline
\multicolumn{1}{c}{GPT-2 \cite{radford2018improving} + RWP + CL}   & \multicolumn{1}{c}{\textbf{88.46}} & \multicolumn{1}{c}{\textbf{74.52}} & \multicolumn{1}{c}{\textbf{75.09}}
\\
\hline
\bottomrule
\end{tabular}
\end{table*}
\endgroup

On 2 out of 3 datasets, DeBERTa \cite{he2020deberta} performed better with noise. However, the performance of DeBERTa \cite{he2020deberta} on the COVID-19 PHM \cite{luo2022covid} dataset decreased with our proposed training method. Similarly, the difference between DeBERTa \cite{he2020deberta}  performance with our proposed method and its corresponding baseline method on the HMC-2019 \cite{biddle2020leveraging} dataset is smaller than the other models. One possible explanation for this behavior is the existence of a built-in virtual adversarial training algorithm in the DeBERTa \cite{he2020deberta}, which already adds perturbation. Since COVID-19 PHM \cite{luo2022covid} is an imbalanced dataset, the performance of the DeBERTa \cite{he2020deberta} degrades on this dataset with additional perturbations. On the other hand, the length of tweets in the HMC-2019 \cite{biddle2020leveraging} dataset is smaller, hence, additional perturbation with our proposed method doesn't help DeBERTa \cite{he2020deberta} much as compared to the other models. However, as the text length of the RHMD \cite{naseem2022identification} dataset is longer, additional perturbations using DeBERTa \cite{he2020deberta} with our proposed training method outperforms its baseline method significantly.

One possible explanation for the improved performance of the models using the proposed training strategy might be the prevention of over-fitting. Generally, models learn and memorize the training set quickly, and are unable to learn contextual and semantic information properly during the training phase. However, perturbations prevent the memorization and encourage the generalization of the models by learning the appropriate representations. CL further improves the representations of the given input text hence improves the models performance.

We further extended the experimentation and performed 10-fold cross-validation for the RHMD \cite{naseem2022identification} and HMC-2019 \cite{biddle2020leveraging} datasets. For these experiments, we used the best validation set hyperparameters of the train/validation/test splits. The results are presented in Table \ref{tbl:sota}. Results suggested that different models exhibit different performance characteristics across 2 datasets. This motivated us to propose a meta predictor that takes ensemble predictions from these models. The results are presented in the section \ref{ens}.  
\begingroup
\setlength{\tabcolsep}{2pt} % Default value: 6pt
\renewcommand{\arraystretch}{1.3}
\begin{table*}[!htbp]
\begin{center}
\caption{Results (macro F1-score) on 10-fold cross-validation set for the RHMD \cite{naseem2022identification} and HMC-2019 \cite{biddle2020leveraging} datasets.}

\label{tbl:sota}
\begin{tabular}{lll}
\toprule
\hline
\multicolumn{1}{c}{Method}      & \multicolumn{1}{c}{HMC-2019 \cite{biddle2020leveraging}}        & \multicolumn{1}{c}{RHMD \cite{naseem2022identification}} \\
\hline
\multicolumn{1}{c}{BERT \cite{devlin2018bert} + RWP + CL}  & \multicolumn{1}{c}{93.62}    & \multicolumn{1}{c}{82.16}   \\
\hline
\multicolumn{1}{c}{RoBERTa \cite{liu2019roberta} + RWP + CL} & \multicolumn{1}{c}{94.27}   & \multicolumn{1}{c}{83.05}   \\
\hline
\multicolumn{1}{c}{DeBERTa \cite{he2020deberta} + RWP + CL} & \multicolumn{1}{c}{94.07} & \multicolumn{1}{c}{82.85}     \\
\hline
\multicolumn{1}{c}{XLNet \cite{yang2019xlnet} + RWP + CL} & \multicolumn{1}{c}{93.47} & \multicolumn{1}{c}{81.91}     \\
\hline
\multicolumn{1}{c}{GPT-2 \cite{radford2018improving} + RWP + CL} & \multicolumn{1}{c}{89.61}  & \multicolumn{1}{c}{74.70}    \\
\hline
\bottomrule
\end{tabular}

\end{center}
\end{table*}
\endgroup

\subsection{Proposed Meta Predictor Performance Analysis and Comparison with SotA HMC approaches}\label{ens}

\begin{figure*}

\centering

\begin{subfigure}[b]{.49\linewidth}
\includegraphics[width=\linewidth]{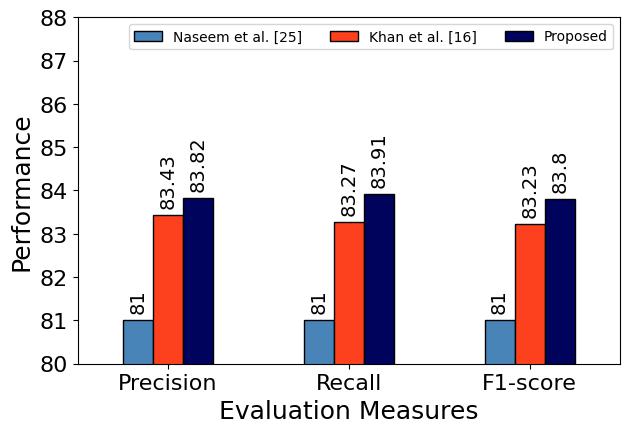}
\caption{RHMD dataset \cite{naseem2022identification}.}
  \label{fig:rhmd-sota}
\end{subfigure}
\begin{subfigure}[b]{.49\linewidth}
\includegraphics[width=\linewidth]{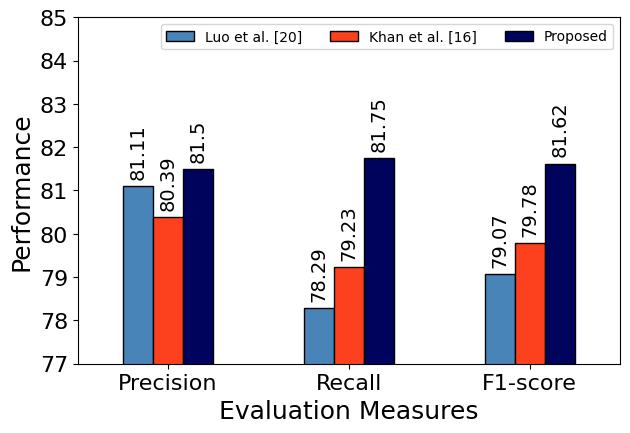}
\caption{COVID-19 PHM dataset \cite{luo2022covid}.}
  \label{fig:covid-19-sota}
\end{subfigure}

\begin{subfigure}[b]{.95\linewidth}
\includegraphics[width=\linewidth]{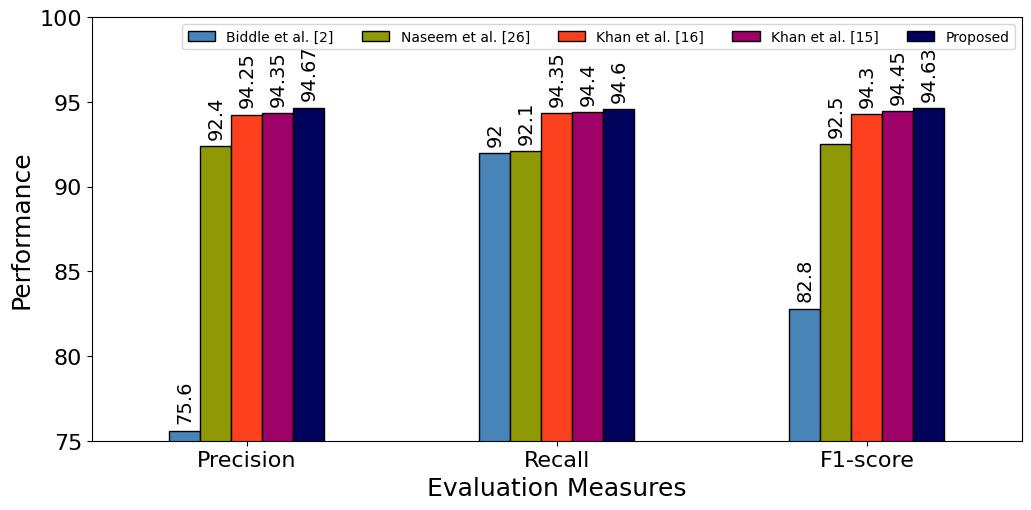}
\caption{HMC-2019 dataet \cite{biddle2020leveraging}.}
\label{fig:hmc-sota}
\end{subfigure}
\caption{Performance comparison of the proposed meta predictor built on the top of proposed training strategy with SotA on 3 datasets. Results of the SotA are taken from the Khan et al. \cite{khan2022improving}. For fair comparison with existing predictors, RHMD \cite{naseem2022identification} and HMC-2019 \cite{biddle2020leveraging} datasets are evaluated using a 10-fold cross-validation.}
\label{fig:sota}
\end{figure*}
Fig. \ref{fig:sota} presents the comparison of the proposed meta predictor with existing predictors. Among the existing predictors, Khan et al. \cite{khan2022improving} produced better performance on the RHMD \cite{naseem2022identification}, COVID-19 PHM datasets \cite{luo2022covid}, and Khan et al. \cite{khan2022novel} on the HMC-2019 \cite{biddle2020leveraging} dataset due to addition of perturbations in the embedding matrix, and representation space. However, like Luo et al.\cite{luo2022covid} method, these methods are biased towards precision on COVID-19 PHM \cite{luo2022covid} dataset. The proposed predictor reduced the difference between precision and recall by utilizing RWP, and benefiting from the capabilities of 5 different models. Overall, our proposed meta predictor achieves the SotA performance on all 3 datasets.
\subsection{Ablation Studies}\label{abl}
\begingroup
\setlength{\tabcolsep}{3pt} % Default value: 6pt
\renewcommand{\arraystretch}{1.3}
\begin{table*}[!htbp]\centering
\caption{showing the effectiveness of our proposed method. Results are reported in terms of macro F1-score.} 
\label{tbl:abl}
\begin{tabular}{lllll}
\toprule
\hline
\multicolumn{1}{c}{Dataset}      & \multicolumn{1}{c}{Model}   & \multicolumn{1}{c}{Baseline} & \multicolumn{1}{c}{RWP}   & \multicolumn{1}{c}{RWP + CL} \\
\toprule
\hline
\multicolumn{1}{c}{HMC-2019 \cite{biddle2020leveraging}}    & \multicolumn{1}{c}{XLNet \cite{yang2019xlnet}}   & \multicolumn{1}{c}{92.53}    & \multicolumn{1}{c}{93.03} & \multicolumn{1}{c}{93.28}    \\
\hline
\multicolumn{1}{c}{RHMD \cite{naseem2022identification}}         & \multicolumn{1}{c}{DeBERTa \cite{he2020deberta}} & \multicolumn{1}{c}{81.03}    & \multicolumn{1}{c}{80.31}   & \multicolumn{1}{c}{82.36}    \\
\hline
\multicolumn{1}{c}{COVID-19 PHM \cite{luo2022covid}} & \multicolumn{1}{c}{GPT-2 \cite{radford2018improving}}   & \multicolumn{1}{c}{71.22}    & \multicolumn{1}{c}{71.59} & \multicolumn{1}{c}{75.09}   \\
\hline
\bottomrule
\end{tabular}
\end{table*}
\endgroup
In Table \ref{tbl:abl}, we present the effectiveness of our proposed method by dropping the CL and RWP components. These results indicate that performance of the models decreases if two proposed components are dropped. 

In Table \ref{tbl:noise}, we present the noise impact on the performance of the models. Generally, models performances significantly drops for higher noise values, i.e., $\epsilon= 5e^{-3}$. BERT \cite{devlin2018bert} is more sensitive to noise on RHMD \cite{naseem2022identification} and COVID-19 PHM \cite{luo2022covid} datasets. The performance of RoBERTa \cite{liu2019roberta} and GPT-2 \cite{radford2018improving} varies significantly for all the datasets. For HMC-2019 \cite{biddle2020leveraging}, the performance of DeBERTa \cite{he2020deberta} does not change much with change in the noise. The performance of XLNet \cite{yang2019xlnet} changes sharply for the higher values of $\lambda$ on HMC-2019 \cite{biddle2020leveraging}, and RHMD \cite{naseem2022identification} datasets.
% \usepackage{multirow}
\begin{table*}[]
\centering
\caption{showing the amount of noise impact on validation sets performance (macro F1-score) of 3 datasets.}
\begin{adjustbox}{width=1.0\textwidth,center=\textwidth}
\label{tbl:noise}
\begin{tabular}{lp{7mm}lllllllllllll}
\toprule
\toprule
\multirow{3}{*}{Model}   & \multicolumn{1}{c}{\multirow{3}{*}{$\lambda$}} & \multicolumn{4}{c}{HMC-2019\cite{biddle2020leveraging}}    & \multicolumn{4}{c}{RHMD \cite{naseem2022identification}}        & \multicolumn{4}{c}{COVID-19 PHM \cite{luo2022covid}} \\
\cmidrule(rl){3-14}
\multicolumn{14}{c}{Noise Amount $\epsilon$} \\
 \cmidrule(rl){3-6} \cmidrule(rl){7-10} \cmidrule(rl){11-14}

                         & \multicolumn{1}{c}{}                       & |  $1e^{-4}$ |  & $5e^{-4}$ |  & $1e^{-3}$ |  & $5e^{-3}$ |  & $1e^{-4}$ |  & $5e^{-4}$ |  & $1e^{-3}$ | & $5e^{-3}$ |  & $1e^{-4}$ |  & $5e^{-4}$ |  & $1e^{-3}$ & | $5e^{-3}$ | \\
                          \cmidrule(rl){3-6} \cmidrule(rl){7-10} \cmidrule(rl){11-14}
\multirow{5}{*}{ \begin{turn}{45} BERT  \cite{devlin2018bert} \end{turn} }   & \multicolumn{1}{c}{0.1}                                        & \multicolumn{1}{c}{93.75}  & \multicolumn{1}{c}{93.39}  & \multicolumn{1}{c}{93.72} & \multicolumn{1}{c}{92.60} & \multicolumn{1}{c}{81.20}  & \multicolumn{1}{c}{81.62}  & \multicolumn{1}{c}{81.26} & \multicolumn{1}{c}{81.60} & \multicolumn{1}{c}{77.86}   & \multicolumn{1}{c}{78.24}  & \multicolumn{1}{c}{78.57} & \multicolumn{1}{c}{78.31} \\
                         & \multicolumn{1}{c}{0.2}                                        & \multicolumn{1}{c}{92.99}  & \multicolumn{1}{c}{92.91}  & \multicolumn{1}{c}{93.27} & \multicolumn{1}{c}{93.25} & \multicolumn{1}{c}{81.42}  & \multicolumn{1}{c}{80.69}  & \multicolumn{1}{c}{81.22} & \multicolumn{1}{c}{82.13} & \multicolumn{1}{c}{78.41}   & \multicolumn{1}{c}{78.81}  & \multicolumn{1}{c}{77.60} & \multicolumn{1}{c}{79.16} \\
                         & \multicolumn{1}{c}{0.3}                                        & \multicolumn{1}{c}{93.17}  & \multicolumn{1}{c}{93.04}  & \multicolumn{1}{c}{93.36} & \multicolumn{1}{c}{93.01} & \multicolumn{1}{c}{80.60}  & \multicolumn{1}{c}{81.58}  & \multicolumn{1}{c}{80.93} & \multicolumn{1}{c}{81.98} & \multicolumn{1}{c}{77.60}   & \multicolumn{1}{c}{78.42}  & \multicolumn{1}{c}{78.42} & \multicolumn{1}{c}{77.76} \\
                         & \multicolumn{1}{c}{0.4}                                        & \multicolumn{1}{c}{93.20}  & \multicolumn{1}{c}{93.17}  & \multicolumn{1}{c}{93.46} & \multicolumn{1}{c}{92.78} & \multicolumn{1}{c}{80.97}  & \multicolumn{1}{c}{80.75}  & \multicolumn{1}{c}{82.05} & \multicolumn{1}{c}{81.74} & \multicolumn{1}{c}{77.80}   & \multicolumn{1}{c}{77.54}  & \multicolumn{1}{c}{77.32} & \multicolumn{1}{c}{78.14} \\
                         & \multicolumn{1}{c}{0.5 }                                       & \multicolumn{1}{c}{93.31}  & \multicolumn{1}{c}{92.98}  & \multicolumn{1}{c}{93.31} & \multicolumn{1}{c}{93.12} & \multicolumn{1}{c}{81.21}  & \multicolumn{1}{c}{81.30}  & \multicolumn{1}{c}{80.46} & \multicolumn{1}{c}{81.85} & \multicolumn{1}{c}{77.00}   & \multicolumn{1}{c}{76.84}  & \multicolumn{1}{c}{76.53} & \multicolumn{1}{c}{77.26} \\
                         \hline
                         \midrule
\multirow{5}{*}{ \begin{turn}{45} RoBERTa \cite{liu2019roberta} \end{turn}} & \multicolumn{1}{c}{0.1}                                        & \multicolumn{1}{c}{94.48}  & \multicolumn{1}{c}{93.94}  & \multicolumn{1}{c}{93.75} & \multicolumn{1}{c}{93.27} & \multicolumn{1}{c}{82.99}  & \multicolumn{1}{c}{83.85}  & \multicolumn{1}{c}{83.08} & \multicolumn{1}{c}{83.33} & \multicolumn{1}{c}{80.12}   & \multicolumn{1}{c}{79.31}  & \multicolumn{1}{c}{80.14} & \multicolumn{1}{c}{78.44} \\
                         & \multicolumn{1}{c}{0.2}                                        & \multicolumn{1}{c}{93.94}  & \multicolumn{1}{c}{94.26}  & \multicolumn{1}{c}{93.97} & \multicolumn{1}{c}{92.91} & \multicolumn{1}{c}{82.01}  & \multicolumn{1}{c}{83.40}  & \multicolumn{1}{c}{83.64} & \multicolumn{1}{c}{83.12} & \multicolumn{1}{c}{79.58}   & \multicolumn{1}{c}{79.31}  & \multicolumn{1}{c}{79.86} & \multicolumn{1}{c}{78.08} \\
                         & \multicolumn{1}{c}{0.3}                                        & \multicolumn{1}{c}{94.04}  & \multicolumn{1}{c}{94.15}  & \multicolumn{1}{c}{94.40} & \multicolumn{1}{c}{92.49} & \multicolumn{1}{c}{82.61}  & \multicolumn{1}{c}{83.77}  & \multicolumn{1}{c}{83.30} & \multicolumn{1}{c}{82.40} & \multicolumn{1}{c}{79.99}   & \multicolumn{1}{c}{79.97}  & \multicolumn{1}{c}{79.23} & \multicolumn{1}{c}{77.27} \\
                         & \multicolumn{1}{c}{0.4}                                        & \multicolumn{1}{c}{94.32}  & \multicolumn{1}{c}{93.75}  & \multicolumn{1}{c}{94.21} & \multicolumn{1}{c}{92.67} & \multicolumn{1}{c}{83.10}  & \multicolumn{1}{c}{83.15}  & \multicolumn{1}{c}{83.17} & \multicolumn{1}{c}{81.26} & \multicolumn{1}{c}{79.71}   & \multicolumn{1}{c}{79.31}  & \multicolumn{1}{c}{79.03} & \multicolumn{1}{c}{61.32} \\
                         & \multicolumn{1}{c}{0.5}                                        & \multicolumn{1}{c}{94.34}  & \multicolumn{1}{c}{94.10}  & \multicolumn{1}{c}{93.71} & \multicolumn{1}{c}{91.72} & \multicolumn{1}{c}{82.91}  & \multicolumn{1}{c}{82.72}  & \multicolumn{1}{c}{82.70} & \multicolumn{1}{c}{82.17} & \multicolumn{1}{c}{78.70}   & \multicolumn{1}{c}{78.61}  & \multicolumn{1}{c}{78.69} & \multicolumn{1}{c}{76.29} \\
                         \hline
                         \midrule
\multirow{5}{*}{\begin{turn}{45} DeBERTa \cite{he2020deberta} \end{turn} } & \multicolumn{1}{c}{0.1}                                        & \multicolumn{1}{c}{94.08}  & \multicolumn{1}{c}{93.89}  & \multicolumn{1}{c}{93.84} & \multicolumn{1}{c}{93.68} & \multicolumn{1}{c}{83.29}  & \multicolumn{1}{c}{83.81}  & \multicolumn{1}{c}{83.51} & \multicolumn{1}{c}{83.58} & \multicolumn{1}{c}{80.44}   & \multicolumn{1}{c}{81.41}  & \multicolumn{1}{c}{80.09} & \multicolumn{1}{c}{81.33} \\
                         & \multicolumn{1}{c}{0.2}                                        & \multicolumn{1}{c}{94.13}  & \multicolumn{1}{c}{94.12}  & \multicolumn{1}{c}{93.96} & \multicolumn{1}{c}{93.91} & \multicolumn{1}{c}{83.67}  & \multicolumn{1}{c}{83.17}  & \multicolumn{1}{c}{83.03} & \multicolumn{1}{c}{83.87} & \multicolumn{1}{c}{80.59}   & \multicolumn{1}{c}{79.94}  & \multicolumn{1}{c}{80.00} & \multicolumn{1}{c}{80.25} \\
                         & \multicolumn{1}{c}{0.3}                                        & \multicolumn{1}{c}{93.98}  & \multicolumn{1}{c}{93.86}  & \multicolumn{1}{c}{94.08} & \multicolumn{1}{c}{93.65} & \multicolumn{1}{c}{82.45}  & \multicolumn{1}{c}{82.83}  & \multicolumn{1}{c}{82.82} & \multicolumn{1}{c}{82.86} & \multicolumn{1}{c}{80.56}   & \multicolumn{1}{c}{80.99}  & \multicolumn{1}{c}{80.53} & \multicolumn{1}{c}{80.19} \\
                         & \multicolumn{1}{c}{0.4}                                        & \multicolumn{1}{c}{94.08}  & \multicolumn{1}{c}{94.03}  & \multicolumn{1}{c}{94.08} & \multicolumn{1}{c}{94.10} & \multicolumn{1}{c}{83.34}  & \multicolumn{1}{c}{82.92}  & \multicolumn{1}{c}{83.52} & \multicolumn{1}{c}{82.97} & \multicolumn{1}{c}{80.34}   & \multicolumn{1}{c}{80.77}  & \multicolumn{1}{c}{80.00} & \multicolumn{1}{c}{79.35} \\
                         & \multicolumn{1}{c}{0.5}                                        & \multicolumn{1}{c}{93.82}  & \multicolumn{1}{c}{93.84}  & \multicolumn{1}{c}{93.65} & \multicolumn{1}{c}{93.69} & \multicolumn{1}{c}{82.70}  & \multicolumn{1}{c}{82.76}  & \multicolumn{1}{c}{82.84} & \multicolumn{1}{c}{82.63} & \multicolumn{1}{c}{79.28}   & \multicolumn{1}{c}{80.34}  & \multicolumn{1}{c}{80.23} & \multicolumn{1}{c}{79.01} \\
                         \hline
                         \midrule
\multirow{5}{*}{\begin{turn}{45} XLNet \cite{yang2019xlnet} \end{turn}}   & 0.1                                        & 93.74  & 93.60  & 93.07 & 92.62 & 81.95  & 82.37  & 82.30 & 80.72 & 79.26   & 78.70  & 78.52 & 77.06 \\
                         & 0.2                                        & 92.65  & 93.28  & 93.23 & 92.13 & 81.44  & 81.99  & 81.81 & 81.56 & 78.35   & 79.83  & 78.03 & 77.32 \\
                         & 0.3                                        & 93.22  & 93.18  & 93.50 & 92.05 & 82.78  & 82.73  & 82.64 & 80.40 & 79.07   & 77.58  & 78.43 & 77.36 \\
                         & 0.4                                        & 92.72  & 93.55  & 93.63 & 92.59 & 82.03  & 80.94  & 81.75 & 78.47 & 79.34   & 77.76  & 77.97 & 73.96 \\
                         & 0.5                                        & 93.37  & 93.37  & 93.32 & 91.09 & 81.58  & 80.67  & 80.97 & 79.48 & 77.98   & 79.41  & 78.59 & 75.81 \\
                         \hline
                         \midrule
\multirow{5}{*}{\begin{turn}{45}GPT-2 \cite{radford2018improving} \end{turn}}   & 0.1                                        & 89.85  & 89.92  & 89.67 & 87.99 & 75.15  & 74.52  & 77.20 & 71.28 & 74.08   & 74.94  & 74.25 & 73.09 \\
                         & 0.2                                        & 88.19  & 89.52  & 89.33 & 86.92 & 74.16  & 72.46  & 75.21 & 72.13 & 74.15   & 74.27  & 73.29 & 73.20 \\
                         & 0.3                                        & 90.75  & 90.15  & 89.45 & 88.42 & 74.58  & 75.69  & 75.54 & 71.91 & 73.83   & 75.26  & 74.61 & 73.20 \\
                         & 0.4                                        & 89.46  & 89.75  & 90.35 & 85.55 & 77.50  & 74.90  & 72.83 & 72.39 & 74.61   & 73.51  & 75.37 & 72.68 \\
                         & 0.5                                        & 90.31  & 89.42  & 89.54 & 86.72 & 76.33  & 73.39  & 71.60 & 69.77 & 74.32   & 73.37  & 73.22 & 71.54 \\
                         \hline
                         \bottomrule
                         
\end{tabular}
\end{adjustbox}
\end{table*}
\section{Conclusion}
This paper proposed a robust meta predictor that will improve the healthcare system by monitoring social media content related to health. The proposed predictor employed a novel training strategy that improved the predictive performance of the language models on 3 datasets. The improved training process prevents the model from quickly over-fitting, which allows it to learn semantic features. The proposed meta predictor exploits the strengths of the 5 models and outperforms existing SotA methods on all benchmark datasets. In future work, the performance of each language model may be enhanced by combining different perturbation and contrastive learning methods. In this paper, we used the same perturbation method and a single contrastive learning approach for training these models.


\bibliographystyle{splncs04}
\bibliography{references}
\end{document}